\documentclass[runningheads]{llncs}
\usepackage{makeidx}
\usepackage{graphicx}
\usepackage{amsmath,amssymb} 
\usepackage{color}
\usepackage{fixltx2e}

\usepackage{booktabs}
\usepackage[percent]{overpic}
\usepackage{url}

\newcommand{\reffig}[1]{Fig.~\ref{fig:#1}}
\newcommand{\reftab}[1]{Tab.~\ref{tab:#1}}
\newcommand{\subsubsubsection}[1]{\vspace{4px} \noindent \textbf{#1}}
\newcommand\norm[1]{\left\lVert#1\right\rVert}
\newcommand{\etal}{\textit{et al. }}
\newcommand{\eg}{\textit{e.g. }}
\newcommand{\ie}{\textit{i.e. }}

\newcommand{\rulesep}{\unskip\ \vrule\ }

\begin{document}
	\pagestyle{headings}
	\mainmatter

	\def\GCPR19SubNumber{113}

	\title{3D Bird's-Eye-View Instance Segmentation}

	\titlerunning{3D Birds-Eye-View Instance Segmentation}
	\authorrunning{C. Elich, F. Engelmann, T. Kontogianni, B. Leibe}
	\author{Cathrin Elich\inst{1,2}\orcidID{0000-0002-3269-6976} \and 
		Francis Engelmann\inst{1}\orcidID{0000-0001-5745-2137} \and 
		Theodora Kontogianni\inst{1}\orcidID{0000-0002-8754-8356} \and
	 	\\ Bastian Leibe\inst{1}\orcidID{0000-0003-4225-0051}}
	\institute{$^1$RWTH Technical University Aachen, Germany\\
	$^2$Max Planck Institute for Intelligent Systems, Tuebingen, Germany}

	\maketitle

\begin{abstract}
  Recent deep learning models achieve impressive results on 3D scene analysis tasks by operating directly on unstructured point clouds.
A lot of progress was made in the field of object classification and semantic segmentation.
However, the task of instance segmentation is currently less explored.
In this work, we present 3D-BEVIS (\textit{3D bird's-eye-view instance segmentation}),
a deep learning framework for joint semantic- and instance-segmentation on 3D point clouds.
Following the idea of previous proposal-free instance segmentation approaches,
our model learns a feature embedding and groups the obtained feature space into semantic instances.
Current point-based methods process local sub-parts of a full scene independently, followed by a heuristic merging step.
However, to perform instance segmentation by clustering on a full scene, globally consistent features are required.
Therefore, we propose to combine local point geometry with global context information using an intermediate bird's-eye view representation.
\end{abstract}


\section{Introduction}
\label{sec:introduction}
The recent progress in deep learning techniques along with the rapid availability of commodity 3D sensors \cite{realsense,matterport,fabscan}
has allowed the community to leverage classical tasks such as semantic segmentation and object detection from the 2D image space into the 3D world.
In this work, we tackle the joint task of semantic segmentation and instance segmentation of 3D point clouds.
Specifically, given a 3D reconstruction of a scene in the form of a point cloud,
our goal is not only to estimate a semantic label for each point but also to identify each object's instance.
Progress in this area is interesting to a number of computer vision applications such as automatic scene parsing, robot navigation and virtual or augmented reality.

\begin{figure}[t!]
	\centering
	\includegraphics[width=1.0\linewidth]{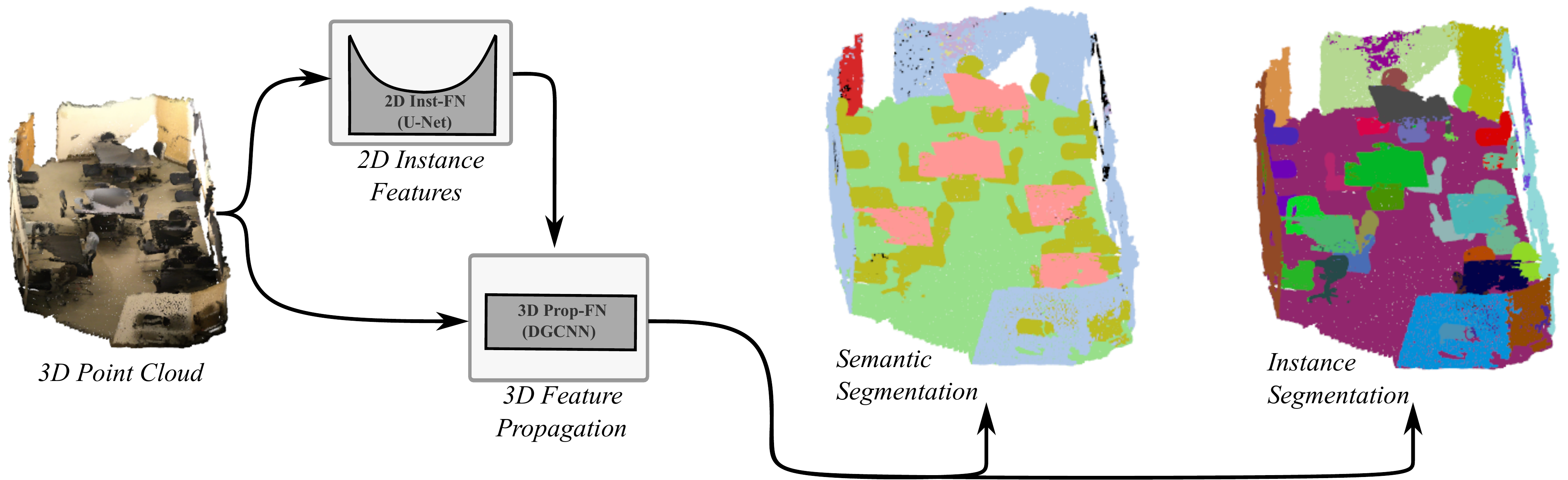}
	\caption{We present a 2D-3D deep model for semantic instance segmentation on 3D point clouds.
	From left to right: The input 3D point cloud, our network architecture combining a 2D U-shaped convolutional network and a 3D graph convolutional network, actual predictions from our method.}
	\label{fig:teaser}
\end{figure}

The main differences between semantic and instance segmentation can be described as follows.
While the semantic segmentation task can be interpreted as a classification task for a fixed number of known labels,
the object indices for instance segmentation are invariant to permutation and the total number of objects is unknown a priori.
Currently, there are two main directions to tackle instance segmentation:
\textit{Proposal-based} methods first look for interesting regions and then segment them into foreground and background \cite{Pinheiro15NIPS,Dai15CoRR,He17ICCV}.
Alternatively, \textit{proposal-free} approaches learn a feature embedding space for the pixels within the image.
The pixels are subsequently grouped according to their feature vector \cite{Newell17NIPS,Hsu18IJCNN,Kong18CVPR}.
In this work, we follow the latter direction since it is straightforward to jointly perform semantic and instance segmentation for every point in the scene.
Moreover, proposal-based approaches generally rely on multi-stage architectures which can be challenging to train.

Two fundamental issues need to be addressed for proposal-free instance segmentation:
First, we need to learn point representations that can be grouped to object instances.
Although some attempts have been made for 2D instance segmentation \cite{Kong18CVPR,Brabandere17CVPRW,Newell17NIPS},
it remains unclear what is the best way to learn instance features on 3D point clouds.
This strongly relates to the second issue which deals with the scale of point clouds.
A typical point cloud can have multiple millions of points along with high dimensional features, including position, color or normals.
The usual approach to deal with large scenes consists in splitting the point cloud into chunks and processing them separately \cite{Qi17CVPR,Qi17NIPS,Wang18CVPR}.
This is problematic for instance segmentation as large instances can extend over multiple chunks.
An alternative is to downsample the original point cloud to a manageable size \cite{Tatarchenko18CVPR} which leads to obvious draw-backs (e.g. loss of detail) and can still fail with very large point clouds, such as in dense outdoor scenes.

In this work, we introduce a hybrid network architecture (see \reffig{teaser}) that learns global instance features on a 2D representation of the full scene and then propagates the learned features onto subsets of the full 3D point cloud.
In order to achieve this effect, we need a network architecture that supports propagation over unstructured data.
The recently presented graph neural network by Wang \etal \cite{Wang18CoRR} for learning a semantic segmentation on point clouds is an adequate choice for this purpose.
We present results for our model on the Stanford Indoor 3D scenes dataset \cite{Armeni16CVPR} and the more recent ScanNet\,v2 dataset \cite{Dai17CVPR}.

The key contributions of this work are as follows:
(1) We present a hybrid 2D-3D network architecture for performing joint semantic and instance segmentation on large scale point clouds.
(2) We show how to combine features learned from a regular 2D representation and unstructured 3D point clouds.


\section{Related Work}
\subsubsubsection{2D Feature Learning for Instance Segmentation.}
Fully convolutional networks (FCN)\,\cite{Shelhamer17PAMI} have been used as part of many successful semantic segmentation methods to provide dense semantic predictions and features \cite{Ronneberger15MICCAI,Badrinarayanan15PAMI,Chen18ECCV}.
Similarly, for proposal-free instance segmentation, pixel-wise features need to be inferred based on which the image pixels can subsequently be clustered.
Fathi \etal \cite{Fathi17CoRR} compute a cross-entropy loss on randomly sampled points for each object instance.
Hsu \etal \cite{Hsu18IJCNN} treat the FCN-features as multinomial distributions.
and rely on the KL-divergence to measure similarities between pixel distributions.
Kong \etal \cite{Kong18CVPR} map pixel embeddings to a hypersphere.
These embeddings are then clustered using a recurrent implementation of the mean-shift algorithm.
Similar to our approach, Brabandere \etal \cite{Brabandere17CVPRW} use a discriminative loss function to penalize large distances between pixels of the same instance and small distances between the mean embeddings of different instances.

While the above approaches are only used for 2D images, we examine instance segmentation on 3D point clouds.
However, our model utilizes an additional 2D representation which has proven to be useful in previous 3D scene understanding tasks \cite{Boulch17CG,Chen17CVPR,Simon18CoRR,Dai18ECCV}.
Building on top of these ideas for 2D feature learning, our model includes a U-shaped \cite{Ronneberger15MICCAI} FCN to process a 2D bird's-eye view learning globally consistent instance features for an entire scene. 

\subsubsubsection{Deep Learning on 3D point clouds.}
Most approaches in 2D vision tasks are taking advantage of powerful features, learned through 2D convolutions.
Extending the use of convolutions to unstructured 3D point cloud data is non-trivial and has become a very active field of research \cite{Qi17CVPR,Qi17NIPS,Tatarchenko18CVPR,Wang18CoRR}.
The seminal work of Qi \etal\,\cite{Qi17CVPR} introduced feature learning directly on raw point clouds through a series of multi-layer perceptrons (\textit{MLPs}) and max-pooling.
Hierarchical features are added in the follow-up work\,\cite{Qi17NIPS}.
In both works, the max-pooling is only able to extract global shape information.
In \emph{dynamic graph CNN} (DGCNN)\,\cite{Wang18CoRR}, PointNets are further generalized by \emph{EdgeConvs} adding local neighborhood information over a k-nearest neighbor graph.
In this work, we rely on DGCNN to learn strong geometric features and simultaneously utilize it as a message passing graph network to propagate learned instance features.

\subsubsubsection{3D Instance Segmentation.}
While recently several approaches were presented for 3D semantic segmentation \cite{Qi17CVPR,Qi17NIPS,Engelmann18ECCVW,Boulch17CG,Wang18CoRR,Engelmann19CoRR,Rethage18ECCV} and object detection \cite{Zhou18CVPR,Qi2017CoRR,Chen17CVPR,Simon18CoRR}, the combined problem of these was mainly disregarded so far.
The only published work that conducts instance segmentation directly on raw 3D point clouds is SGPN\,\cite{Wang18CVPR}.
A pair-wise similarity matrix is computed and subsequently thresholded to generate proposals which are merged according to a confidence score.
As point clouds are split into smaller blocks that are processed separately, a heuristic \emph{GroupMerging} algorithm is required to merge identical instances.
In contrast, the instance features in this work are globally coherent across a scene such that the instances can directly be extracted without the need of a merging algorithm or thresholding.


\section{Model}
 \begin{figure}[t]
	\centering
	\includegraphics[width=1.0\linewidth]{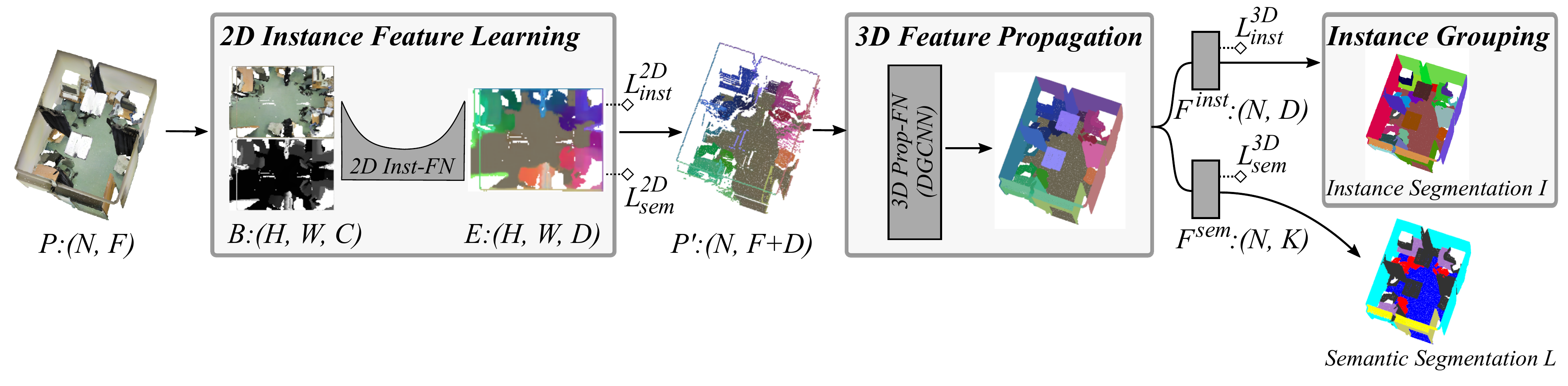}
	\vspace{-9mm}
	\caption{\textbf{3D BEVIS framework.} Given a point cloud $\mathcal{P}$, our model predicts instance labels $\mathcal{I}$ and semantic labels $\mathcal{L}$. The entire pipeline consists of three stages: First, the 2D instance feature network learns instance features $\mathcal{E}$ from a bird's-eye-view $\mathcal{B}$ of the scene. After concatenating the instance features to the original point cloud features, a 3D feature propagation network propagates and predicts instance features for all points in the scene. Our model finally predicts semantic labels $\mathcal{L}$ and instance features $\mathcal{F}^{\text{inst}}$ which are clustered to instance labels $\mathcal{I}$.}
	\label{fig:model}
\end{figure}

In  the following, we will present the architecture of our model 3D-BEVIS for semantic instance segmentation on 3D point clouds as visualized in \reffig{model}.
The input for our model is a point cloud $\mathcal{P} = \{x_i\}_{i=1}^N$, \ie a set of points $x_i \in \mathbb{R}^{F}$ where $F$ is the dimension of the input point features. In our model, we use $F = 9$ for XYZ-position, RGB-color and normalized position with respect to the room size as in \cite{Qi17CVPR}.
The model predicts semantic labels $\mathcal{L}=\{l_i\}_{i=1}^N$ and instance features $\mathcal{F}^{\text{inst}}=\{f_i\}_{i=1}^N$ with $f_i \in \mathbb{R}^{D}$
which are grouped to extract the semantic instance labels
$\mathcal{I} = \{\mathcal{I}_i\}_{i=1}^N$.
The entire framework consists of the combination of a 2D and a 3D feature network to learn point-wise instance features, followed by a clustering procedure to obtain the final instance segmentation.
First, an intermediate 2D representation of the scene is utilized to learn globally consistent instant features for a scattered subset of points. These features are subsequently propagated towards the remaining points of the point cloud by applying a 3D feature propagation network. A clustering with respect to these learned features yields the final objects instances.
Next, the single stages are explained in detail.

\subsection{2D Instance Feature Network.}
\begin{figure}[t]
	\begin{center}
	\includegraphics[width=0.6\linewidth, trim={0 0 0 0}, clip]{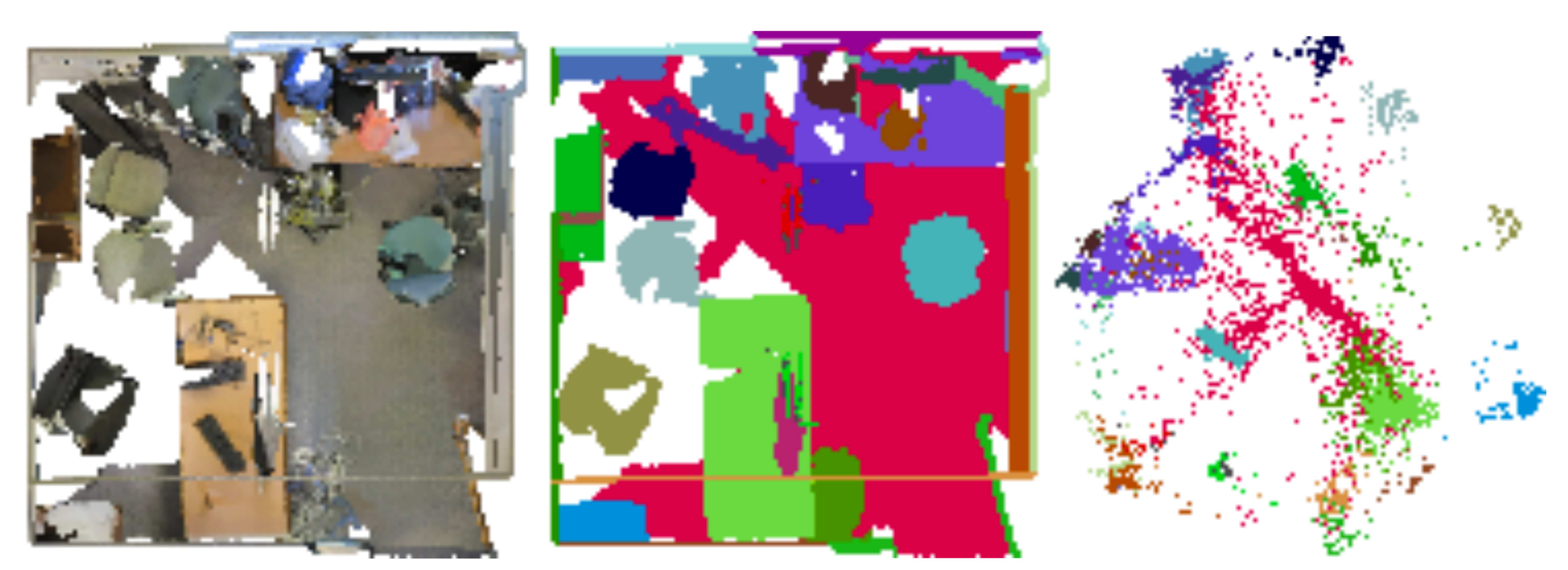}
	\end{center}
	\vspace{-4mm}
	\begin{small}
	\begin{tabular}{ccc}
		\hspace{30mm} Input: $\mathcal{B}$ &
		\hspace{4mm} GT instance seg.  &
		\hspace{4mm} Output: $\mathcal{E}$ \\
	\end{tabular}
	\end{small}
	\vspace{1mm}
	\caption{Left to right: Input bird's-eye view $\mathcal{B}$, ground truth instance labels, predicted instance features $\mathcal{E}$ colored according to the GT instance labels. For visualization, we project the $D$-dimensional instance features $\mathcal{E}$ to 2D with PCA.}
	\label{fig:2d_network_viz}
\end{figure}

To efficiently process the entire scene at once, we consider an intermediate representation $\mathcal{B}\in\mathbb{R}^{H \times W \times C}$ in the form of a bird's-eye view projection of the point cloud $\mathcal{P}$ (see \reffig{2d_network_viz}).
In contrast to previous methods {\cite{Wang18CVPR} that independently process small chunks of the full point cloud, we are thereby able to learn instance features which are globally consistent across the point cloud. 
For generating this view, the points are projected onto a grid on the ground plane. If several points fall into the same cell, only the highest point above the ground plane is taken into account. 
We use color and height-above-ground as input channels, thus $C=4$.
The projections $\mathcal{B}$ are precomputed offline.
The resulting 2D representation is the input to a fully convolutional network (FCN)\,\cite{Shelhamer17PAMI} which predicts the instance feature map $\mathcal{E} \in \mathbb{R}^{H \times W \times D}$.
The FCN can process rooms of changing size during testing.
We utilize a simple encoder-decoder architecture inspired by U-Net \cite{Ronneberger15MICCAI} and the FCN applied in \cite{Hsu18IJCNN}.
Convolutions use a 3x3 kernel size with batch normalization, ReLU non-linearities, and skip-connections.
The full architecture is shown in \reffig{2d_network}.

There are two output branches, one for semantic segmentation and one for instance segmentation.
The corresponding losses are $\mathcal{L}^{2D}_{\text{inst}}$ and $\mathcal{L}^{2D}_{\text{sem}}$.
$\mathcal{L}^{2D}_{sem}$ is the cross-entropy loss for semantic segmentation.
The instance segmentation loss $\mathcal{L}^{2D}_{\text{inst}}$ is based on
a similarity measure for pairs of pixels:
	$s_{i,j}=\norm{x_i-x_j}_2$.
From this, we define the entire loss as 
\begin{equation}
	\mathcal{L}^{2D}_{\text{inst}} = \mathcal{L}_{var} + \mathcal{L}_{dist}
\end{equation}
with
\begin{equation}
\begin{split}
	\mathcal{L}_{var} &= \sum^C_{c=1} \sum_{x_i, x_j\in S_c} [s_{i,j}-\delta_{var}]_+, \\
	\mathcal{L}_{dist} &= \sum^C_{\substack{c, c'=1\\ c\neq c'}} \sum_{\substack{x_i\in S_c\\ x_j \in S_{c'}}} [\delta_{dist}-s_{i,j}]_+
\end{split}
\end{equation}

\begin{figure}[t]
	\centering
	\includegraphics[width=0.8\linewidth]{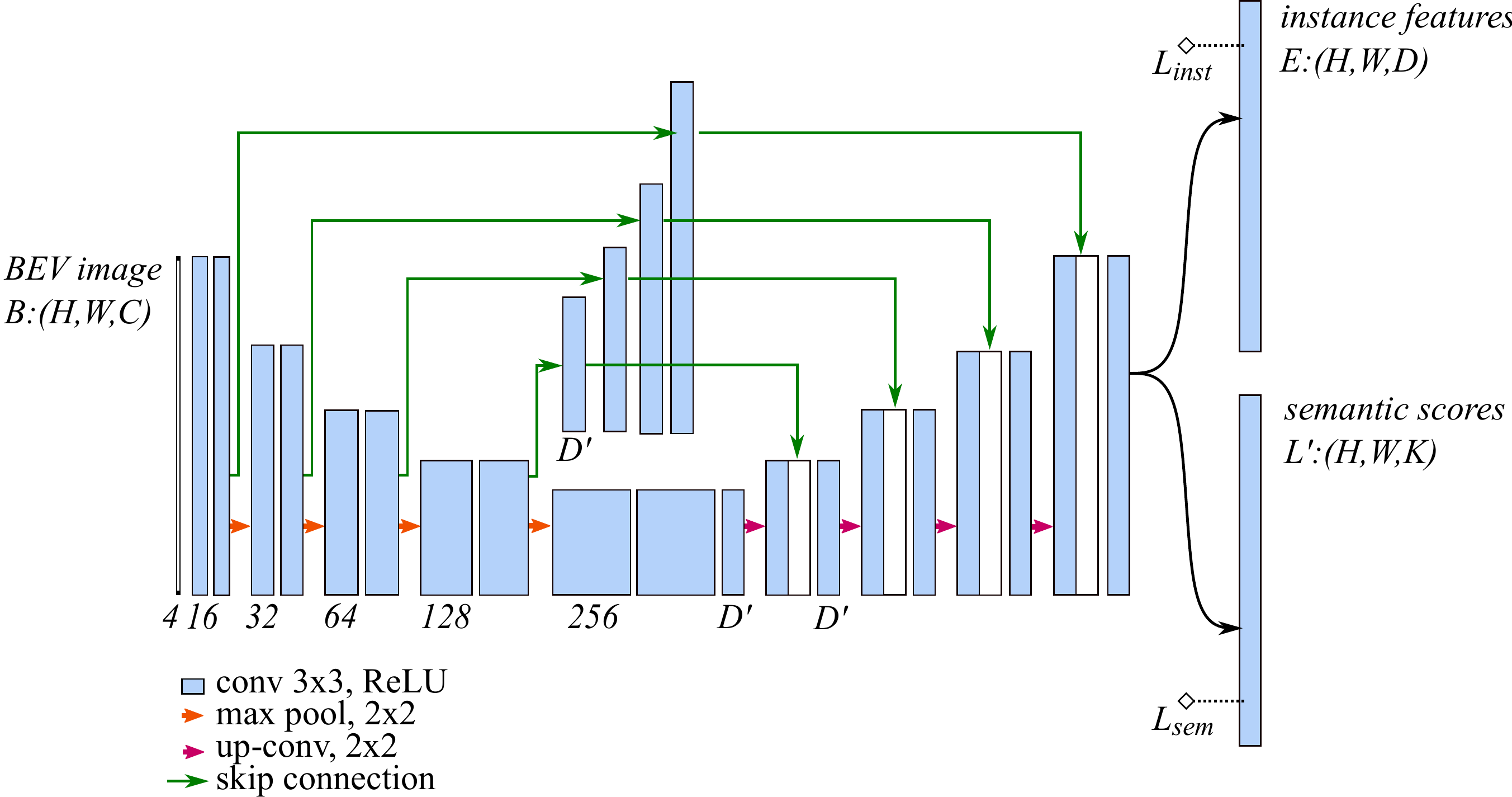}
	\caption{\textbf{2D Instance Feature Network.} U-shaped fully convolution network to learn instances features $\mathcal{E}$ from the input bird's-eye view $\mathcal{B}$.
	During training the network predicts semantic labels and instance features.
	At test time, we only forward the instance features $\mathcal{E}$.}
	\label{fig:2d_network}
\end{figure}

This ensures feature vectors of points belonging to the same object to be similar while encouraging a large distance in the feature space between features corresponding to different instances. 
Whereas the margin $\delta_{var}$ allows instance features to be spread within a certain range, $\delta_{dist}$ enforces a minimum distance between to feature vectors.
$[\cdot]_+$ denotes the hinge function $\max(0,\cdot)$.

To compute the instance loss, we use the same sampling strategy as applied in \cite{Fathi17CoRR,Newell17NIPS}. Instead of comparing all pairs of feature vectors, we sample a subset $S_c$ containing $M$ pixels for each instance $c$.

\subsection{3D Feature Propagation Network.}
At this stage, we have instance features $\mathcal{E}$ for all the points $\mathcal{P}_{\mathcal{B}} \subset \mathcal{P}$ visible in $\mathcal{B}$.
These features are globally consistent and can thus be used as a basis for later grouping. Due to occlusion in the bird's-eye view projection, a fraction of the points was unregarded so far.
Therefore, in this part, we use a graph neural network to propagate existing features and predict instance features for all points in $\mathcal{P}$.
Specifically, we concatenate the initial point cloud features $x_i$ with the learned instance features from $\mathcal{B}$ to obtain $\mathcal{P}'$.
When generating $\mathcal{B}$, we keep track of point indices to map the learned instance features back to the point cloud $\mathcal{P}$.
The instance features of unseen points in $\mathcal{P} \setminus \mathcal{P}_{\mathcal{B}}$ are set to zero.
As graph neural network, we use the architecture from DGCNN \,\cite{Wang18CoRR} which was originally presented for learning a semantic segmentation on point clouds. 
Similar to the 2D instance feature network, the graph neural network has two output branches, each with an assigned loss function. The semantic segmentation loss $\mathcal{L}^{3D}_{sem}$ is again the cross-entropy loss. The instance segmentation loss $\mathcal{L}^{3D}_{inst}$ is defined as:
\begin{equation}
\mathcal{L}^{3D}_{inst} = \norm{\mathcal{F}^{\text{inst}} - \mathcal{F}^{\text{target}}}
\end{equation}
where $\mathcal{F}^{\text{target}} \in \mathbb{R}^{N\times D}$ are \emph{target instance features}.
The target instance feature for a point $x_i$ is the mean over all instance features in $\mathcal{E}$ which lie in the same ground truth instance $\mathcal{I}_j$.
If an instance is not visible in $\mathcal{B}$, there will be no target instance feature. Such instances are not part of the loss during training.

\subsection{Instance Grouping.}
The last component obtains the final instance labels $\mathcal{I}$ by clustering the predicted instance features $\mathcal{F}^{\text{inst}}$ using the MeanShift\,\cite{Comaniciu02PAMI} algorithm.
MeanShift does not require a pre-determined number of clusters and is thus suited for the task of instance segmentation with an arbitrary number of instances.
The semantic labels $\mathcal{L}$ directly correspond to the category with the highest prediction in the semantic output branch of the propagation network.

As a final post-processing step, we found it beneficial to split up instances with an inconsistent semantic labeling. 
More specifically, we obtain a new instance $\mathcal{I}_c$ for every class $c$ if at least $th_c$ points in $\mathcal{I}$ have predicted semantic label $c$. $th_c$ is chosen to be proportional to the average number of points per instance of the respective category.
This helps to distinguish between objects from different classes that are hardly identified from the bird's-eye view like windows and walls.

\subsection{Training details}
We train the 2D instance feature network on bird's-eye-view projections at a resolution of 3\,cm (\textit{S3DIS}) or 5\,cm (\textit{ScanNet}) per pixel. Depending on the room size, images are either cropped or padded.
We deal with ceiling points by heuristically removing the highest points in each point cloud up to a threshold.
As the network is fully convolutional, we can process the full image at test time.
We perform data augmentation on the bird's-eye views $\mathcal{B}$ by random rotation at angles of $90^{\circ}$, scaling and horizontal/vertical flipping.

To optimize the loss of the 3D feature propagation network, we pick a random position and extract 1024 points from a cylindric block with diameter 1\,m$^2$ or 1.5\,m$^2$ on the ground plane. This is comparable to the proceeding in \cite{Qi17CVPR,Wang18CoRR}. The semantic losses are weighted with the negative logarithm of the class frequency.
The networks are trained with the Adam optimizer\,\cite{Kingma15ICLR} using exponential learning rate decay with an initial rate of $10^{-3}$.


\section{Experiments}
We evaluate our method using two benchmark datasets on which we conduct experiments on the task of semantic and instance segmentation.
We show qualitative and quantitative results on both tasks.

\subsection{Settings}
To evaluate our method, we need point cloud datasets with point-wise instance labels and semantic labels for each instance.

\subsubsubsection{Stanford Large-Scale 3D Indoor Spaces (S3DIS)\,\cite{Armeni16CVPR}}
contains dense 3D point clouds from 6 large-scale indoor areas consisting of  271 rooms from 3 different buildings.
The points are annotated with 13 semantic classes and grouped into instances.
We follow the usual 6-fold cross validation strategy for training and testing as used in \cite{Qi17CVPR}.

\subsubsubsection{ScanNet\,v2 \cite{Dai17CVPR}} contains 3D meshes of a wide variety of indoor scenes including apartments, hotels, conference rooms and offices.
The dataset contains 20 semantic classes.
We use the public training, validation and test split of 1201, 312 and 100 scans, respectively.

\subsubsubsection{Metrics.}
For semantic segmentation, we adopt the predominant metrics from the field: intersection over union and overall accuracy.
The overall accuracy is an inadequate measure as it favors classes with many points, as it is also noted in \cite{Tatarchenko18CVPR}.
To report scores on instance segmentation we follow the evaluation scheme applied in \cite{Wang18CVPR} to which we compare.
We report the average precision (AP) of the predicted instances with an overlap of 50\,\% with the ground truth instances
for the single categories as well as the AP with 25\,\% and 75\,\% overlap.
We also report results on the official ScanNet benchmark challenge \cite{Dai17ScannetBenchmark}
which uses a stricter metric that is adapted from the CityScapes\,\cite{Cordts16CVPR} evaluation.
Specifically, this metric penalizes wrong semantic labels even if the instance labels are predicted correctly.
Moreover, false negatives are taken into account for the precision score.

\subsubsubsection{Baselines.}
We compare our method to SGPN\,\cite{Wang18CVPR}, the only published work so far in the field of semantic instance segmentation operating directly on point clouds.
SGPN uses PointNet \cite{Qi17CVPR} as the initial feature extraction network.
We conducted an additional  baseline experiment SGPN$_{\text{DGCNN}}$ which replaces PointNet by DGCNN\,\cite{Wang18CoRR}.
We used the source code provided by the authors of \cite{Wang18CVPR}, although it required some modifications to run.
Due to a lack of information regarding the test split of the dataset used for the provided model, we re-trained the model.
On the ScanNet dataset, we also include the PMRCNN (\textit{Projected MaskRCNN}) baseline experiment provided by the authors of the ScanNet benchmark challenge \cite{Dai17ScannetBenchmark}.
Their method projects predictions on 2D color images into 3D space.

\subsection{Main Results}
\begin{table}[t]
\begin{center}
\begin{tabular}{lccc|cc}
\toprule
	&\multicolumn{3}{c}{Instance Seg.} &\multicolumn{2}{c}{\vspace{-5px}Semantic Seg.}\\
	&\multicolumn{5}{c}{\noindent\rule{0.5\columnwidth}{0.5pt}}\\
     &  AP\textsubscript{0.25}\enspace & \enspace AP\textsubscript{0.5}\enspace & \enspace AP\textsubscript{0.75}\enspace & \enspace mIoU & \enspace mAcc\\
\midrule
SGPN\,\cite{Wang18CVPR} &
 62.47 & 42.91 & 23.89 & \enspace 48.27 & \enspace 71.07 \\
SGPN$_{\text{(DGCNN)}}$ &
70.73 & 58.56 & 39.73 & \enspace \textbf{59.29} & \enspace 80.71\\
\midrule
Ours (3D-BEVIS) &
\textbf{78,45} & \textbf{65,66} & \textbf{46,72}& \enspace 58.37 & \enspace \textbf{83.69} \\
\bottomrule
\end{tabular}
\end{center}
\vspace{-5px}
\caption{\textbf{Instance and semantic segmentation results on the S3DIS\,\cite{Armeni16CVPR} dataset.}
In this table, we compare methods that jointly predict semantic labels and instance labels. 
Our presented method yields the best results for instance segmentation compared to both versions of SGPN.
The semantic scores mainly depend on the 3D feature network and are thus comparable for SGPN$_{\text{(DGCNN)}}$ and 3D-BEVIS.
Using DGCNN as a feature network gives an important improvement.
}
\label{tab:s3dis_summary}
\end{table}

\begin{table}[t]
\begin{center}
\resizebox{\textwidth}{!}{
\begin{tabular}{l|c|cccccccccccc}
\toprule
 & Mean  & ceiling & floor & wall & beam & column & window & door & table & chair & sofa & bookcase & board \\ 
 \midrule
SGPN & 42.90 & 78.15 & 80.27 & 48.90 & 33.65 & 16.97 & 49.63 & 44.48 & 30.33 & 52.22 & 23.12 & 28.50 & 28.62 \\ 
SGPN$_{\text{DGCNN}}$ & 58.56 & \textbf{85.85} & 83.15 & 61.65 & \textbf{52.82} & 47.60 & 55.12 & 62.22 & 34.97 & 66.02 & 42.50 & 55.93 & 54.85 \\  \midrule
3D-BEVIS & \textbf{65.66} & 71.00 & \textbf{96.70} & \textbf{79.37} & 45.10 & \textbf{64.38} & \textbf{64.63} & \textbf{70.15} & \textbf{57.22} & \textbf{74.22} & \textbf{47.92} & \textbf{57.97} & \textbf{59.27} \\ 
\bottomrule
\end{tabular}}
\end{center}
\vspace{-5px}
\caption{\textbf{Category-wise AP\textsubscript{0.5} on S3DIS.} We receive the best results in nearly all categories.}
\label{tab:s3dis_category}
\end{table}

We present quantitative and qualitative results for semantic instance segmentation.
\reftab{s3dis_summary} summarizes our results on S3DIS. Category-wise scores for AP\,50\% are presented in \reftab{s3dis_category}.
Our model outperforms both versions of SGPN over all overlap thresholds and most categories. The relatively low result for the category \textit{ceiling} is due to omitting the ceiling in the bird's eye view. Therefore, the distinction of several such elements is never learned.
We see that DGCNN is a powerful method, it can help to significantly improve the existing approach regarding both the instance and semantic segmentation. 
Please note that our scores differ from the ones reported in SGPN\,\cite{Wang18CVPR}. The difficulty of reproducibility might be due to the considerable number of heuristic thresholds. 

We present detailed results on ScanNet for AP\,50\% in \reftab{scannet_category}. 
In \reftab{scannet_benchmark}, we report our scores on the ScanNet\,v2 benchmark 3D instance segmentation challenge. We get decent results compared to our baseline SGPN. Other recently submitted scores are included as well. Hou \etal \cite{Hou18CoRR} use multi-view RGB-D images as additional input. Yi \etal \cite{Yi18CoRR} predict object proposals on point clouds.
\newline

\begin{table}[t]
\begin{center}
\resizebox{\textwidth}{!}{
\begin{tabular}{l|c|cccccccccccccccccccc}
\toprule
&Mean & wall & floor & cabi- & bed & chair & sofa & table & door & win- & book- & pic- &coun- & desk & cur- & fridge & shower & toilet & sink & bath- & other \\
& & & & net & & & & & & dow &  & ture &ter &  & tain &  & curtain & & &tub & furniture \\ \midrule
SGPN* \cite{Wang18CVPR} & 35.09 & 46.90 & 79.00 & \textbf{34.10} & 43.80 & 63.60 & 36.80 & 40.70 & 0.00 & 0.00 & 22.40 & 0.00 & \textbf{26.90} & 22.80 & \textbf{61.10} & 24.50 & 21.70 & 60.50 & 35.80 & 46.20 & - \\ 
3D-BEVIS  & \textbf{57.73} & \textbf{70.30} & \textbf{97.00} & 29.70 & \textbf{78.30} & \textbf{75.60} & \textbf{65.00} & \textbf{68.50} & \textbf{36.80} & \textbf{37.40} & \textbf{65.00} & \textbf{21.30} & 14.50 & \textbf{37.50} & 57.80 & \textbf{71.40} & \textbf{56.40} & \textbf{68.10} & \textbf{57.40} & \textbf{88.90} & 38.80\\
\bottomrule
\end{tabular}}
\end{center}
\vspace{-5px}
\caption{\textbf{Category-wise AP\textsubscript{0.5} on ScanNet.} The presented scores for SGPN are extracted from \cite{Wang18CVPR}. As they did not provide scores for \textit{other furniture}, this category does not contribute to the mean score.}
\label{tab:scannet_category}
\end{table}

\begin{table}[t]
\begin{center}
\begin{tabular}{lccc}
\toprule
& AP & AP\textsubscript{0.5} & AP\textsubscript{0.25} \\
\midrule
PMRCCN \cite{Dai17ScannetBenchmark}	& 2.1 & 5.3 & 22.7 \\
SGPN \cite{Wang18CVPR} & 4.9  & 14.3  & 39.0\\
Our method	&	11.0 & 22.5 & 35.0\\
\midrule
3D-SIS* \cite{Hou18CoRR} &16.1 & 38.2  &55.8 \\
GSPN* \cite{Yi18CoRR} & 15.8& 30.6 & 54.4\\
\bottomrule
\end{tabular}
\end{center}
\vspace{-5px}
\caption{\textbf{ScanNet\,v2 Benchmark Challenge.}
We report the mean average precision AP at overlap 25\% (AP\textsubscript{0.25}), overlap 50\% (AP\textsubscript{0.5}) and for overlaps in the range [0.5,0.95] with step size 0.05 (AP). 
We report additional submitted scores from concurrent work that was recently accepted for publication (*). 
Scores from \cite{Dai17ScannetBenchmark}.
}
\label{tab:scannet_benchmark}
\end{table}

We show qualitative results of our method for instance and semantic segmentation on S3DIS\,\cite{Armeni16CVPR} in \reffig{s3dis_qualitative_results} and ScanNet\,\cite{Dai17CVPR} in \reffig{scannet_qualitative_results} at the end this paper. Our model can successfully distinguish between several objects of the same category as can be seen \eg regarding multiple chairs within a scene. Visualized inferred features for the 2D bird's eye views are depicted in \reffig{s3dis_2d}.

\begin{figure}
\centering
\begin{minipage}[t]{.37\linewidth}
\centering
\includegraphics[width=\linewidth]{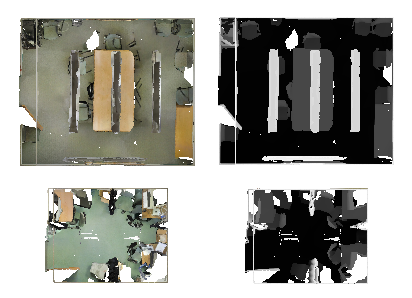}
Input \\
RGB \hspace{11mm} depth
\end{minipage}
    \rulesep
\begin{minipage}[t]{.37\linewidth}
\centering
\includegraphics[width=\linewidth]{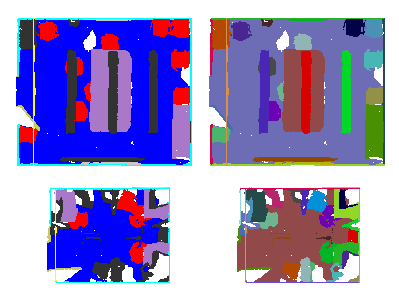}
Segmentation (GT) \\
semantic \hspace{7mm}  instance
\end{minipage}
    \rulesep
\begin{minipage}[t]{.2\linewidth}
\centering
\includegraphics[width=\linewidth]{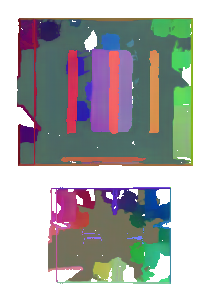}
Inst. features
\end{minipage}
    \caption{\textbf{Predicted instance features for 2D BEV.} Left to right: Input RGB and depth images, ground truth semantic and instance segmentation, instance features. Instance features are mapped into RGB space by applying PCA.}
      \label{fig:s3dis_2d}
\end{figure}

\begin{figure}
\centering
\begin{minipage}[t]{\linewidth}
\centering
\includegraphics[width=0.7\linewidth]{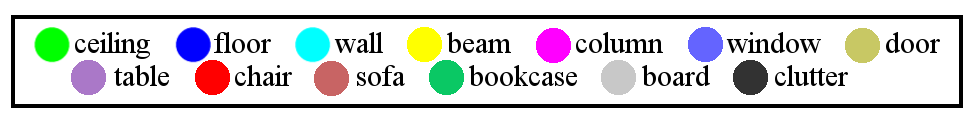}
\end{minipage}\\
\begin{minipage}[t]{.185\linewidth}
\centering
\includegraphics[width=\linewidth]{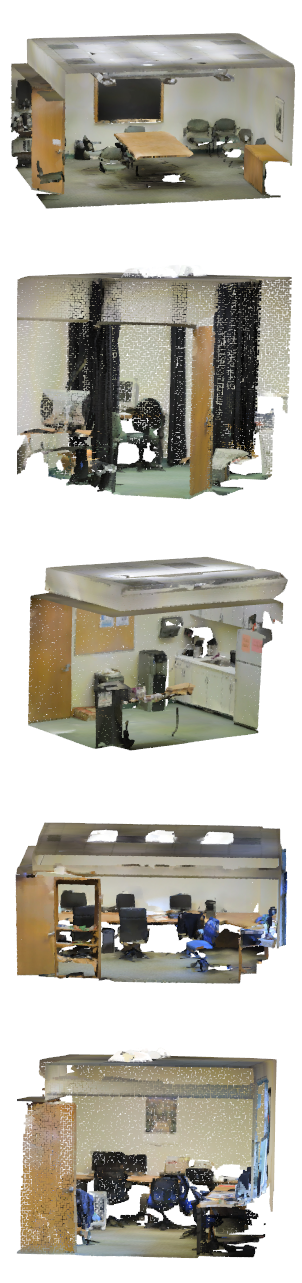}
RGBD Input
\end{minipage}
    \rulesep
\begin{minipage}[t]{.37\linewidth}
\centering
\includegraphics[width=\linewidth]{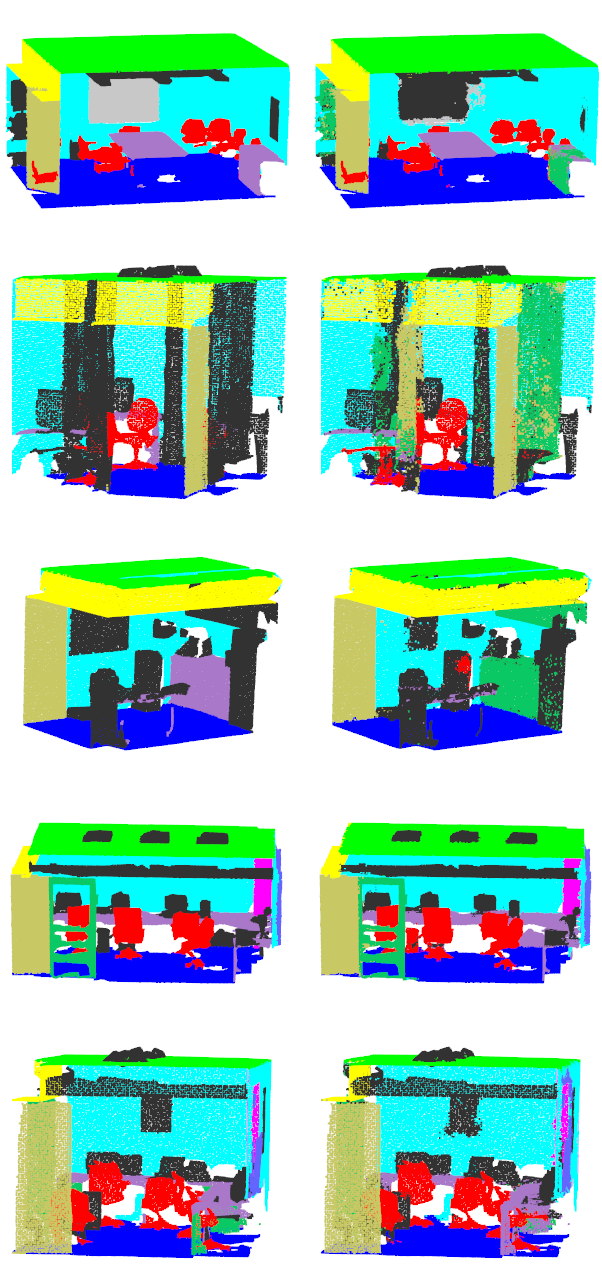}
Semantic Segmentation \\
GT \hspace{10mm} pred.
\end{minipage}
    \rulesep
\begin{minipage}[t]{.37\linewidth}
\centering
\includegraphics[width=\linewidth]{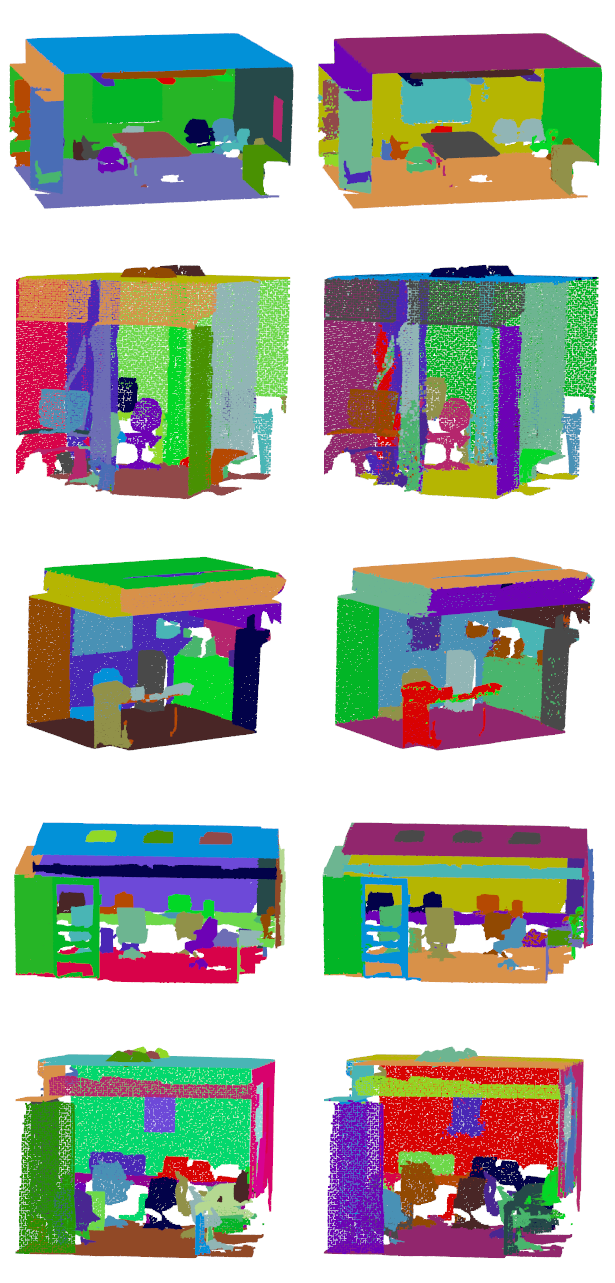}
Instance Segmentation \\
GT \hspace{10mm} pred.
\end{minipage}
    \caption{\textbf{Qualitative results on S3DIS \cite{Armeni16CVPR}.} Left to right: Input RGB point cloud, semantic segmentation (ground truth, prediction), instance segmentation (ground truth, prediction). While we have a fixed color for each class, the color mapping for the single instances is arbitrary.}
      \label{fig:s3dis_qualitative_results}
\end{figure}

\begin{figure}[t!]
	\centering
	\begin{minipage}[t]{\linewidth}
	\centering
\includegraphics[width=0.9\linewidth]{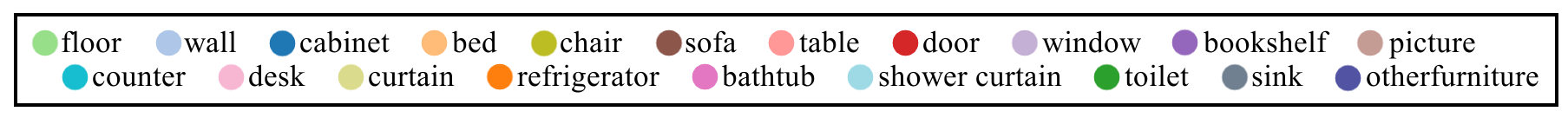}
\end{minipage}\\
\begin{minipage}[t]{.15\linewidth}
\centering
\includegraphics[width=\linewidth]{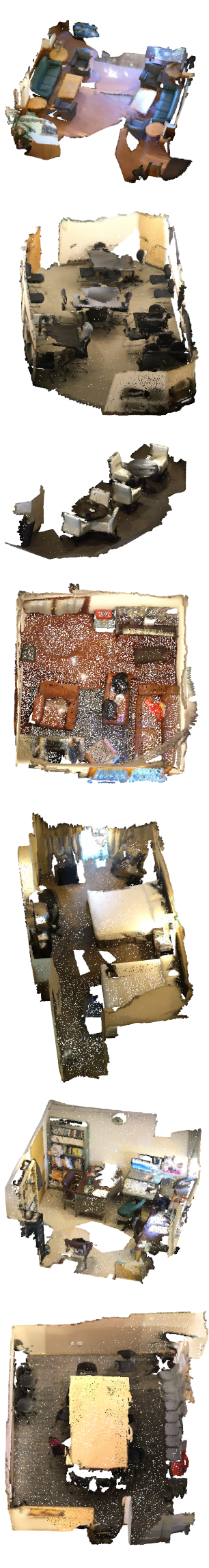}
RGBD Input
\end{minipage}
    \rulesep
\begin{minipage}[t]{.3\linewidth}
\centering
\includegraphics[width=\linewidth]{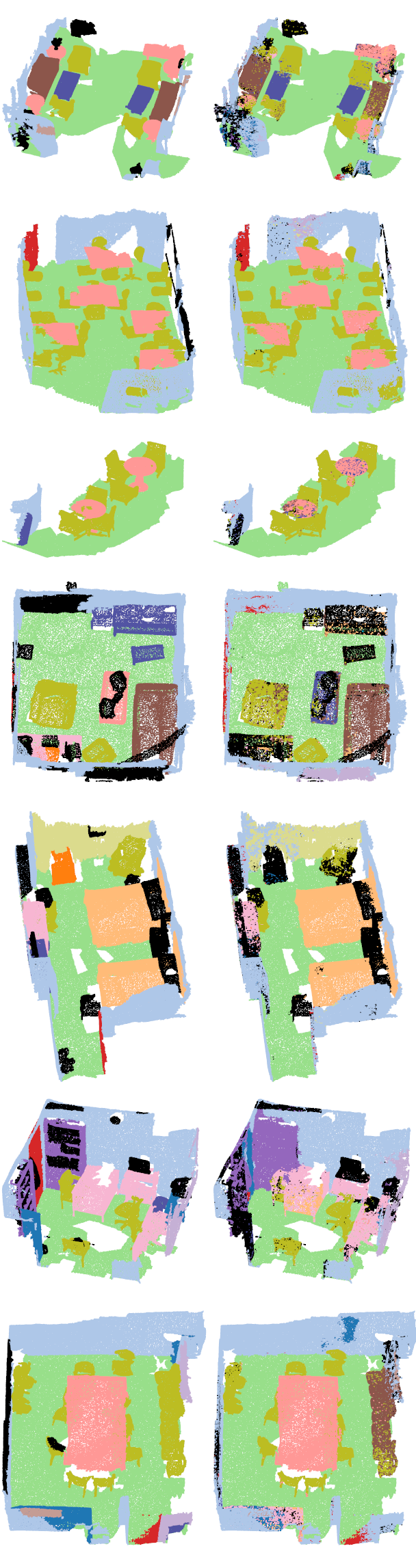}
Semantic Segmentation \\
GT \hspace{10mm} pred.
\end{minipage}
    \rulesep
\begin{minipage}[t]{.3\linewidth}
\centering
\includegraphics[width=\linewidth]{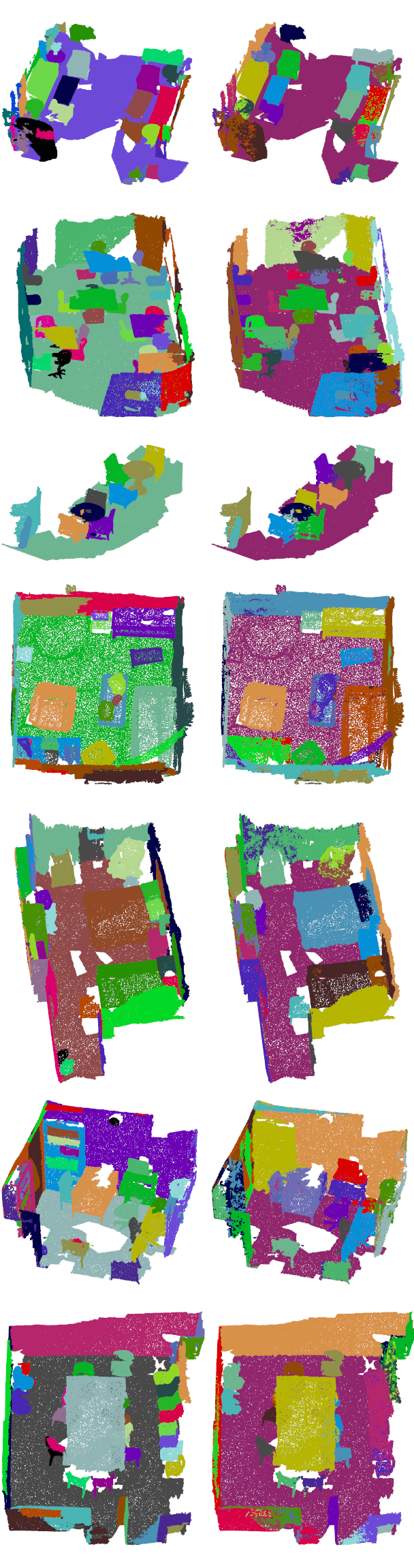}
Instance Segmentation \\
GT \hspace{10mm} pred.
\end{minipage}
    \rulesep
\begin{minipage}[t]{.15\linewidth}
\centering
\includegraphics[width=\linewidth]{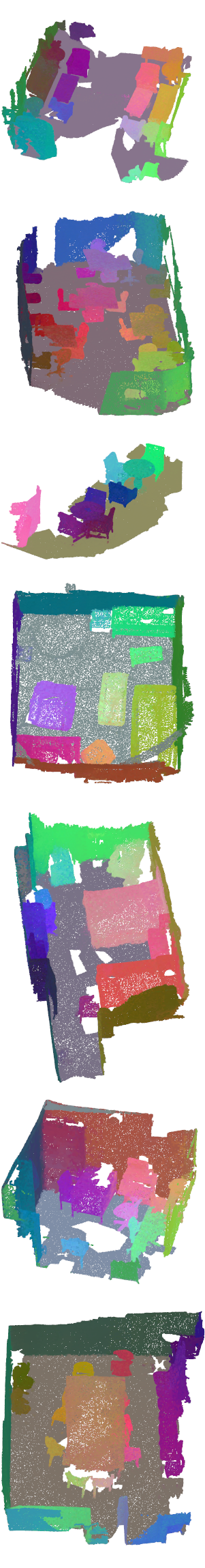}
Instance features
\end{minipage}
	\caption{\textbf{Qualitative results on ScanNet \cite{Dai17CVPR}.} Left to right: Input RGB point cloud, semantic segmentation (ground truth, prediction), instance segmentation (ground truth, prediction). While we have a fixed color for each class, the color mapping for the single instances is arbitrary.}
	\label{fig:scannet_qualitative_results}
\end{figure}

\section{Discussion and Conclusion}
The bird's-eye view used in this work has proven to be very powerful to compute globally consistent features.
However, there are intrinsic limitations, \eg vertically oriented objects are not well visible in this 2D representation.
The same is true for scenes including numerous occluded objects.
An obvious extension could be to include multiple 2D views of the scene.
Compared to previous work\,\cite{Wang18CVPR}, our model is able to learn global instance features which are consistent over a full scene.
Thus, the presented method overcomes the necessity for a heuristic post-processing step to merge instances.

In this work, we explored the relatively new field of instance segmentation on 3D point clouds.
We have proposed a 2D-3D deep learning framework combining a U-shaped fully convolution network to learn globally consistent instance features from a bird's-eye view in combination with a graph neural network to propagate and predict point features in the 3D point cloud.
Future work could look at alternative 2D representations to overcome the limitations of the bird's-eye view.

\clearpage

\bibliographystyle{splncs04}
\bibliography{ms}

\end{document}